\documentclass[sigconf]{acmart}
\usepackage{hyperref}
\AtBeginDocument{%
  \providecommand\BibTeX{{%
    \normalfont B\kern-0.5em{\scshape i\kern-0.25em b}\kern-0.8em\TeX}}}

\setcopyright{acmcopyright}
\copyrightyear{2026}
\acmYear{2026}
\setcopyright{cc}
\setcctype{by}
\acmConference[WWW '26]{Proceedings of the ACM Web Conference 2026}{April 13--17, 2026}{Dubai, United Arab Emirates}
\acmBooktitle{Proceedings of the ACM Web Conference 2026 (WWW '26), April 13--17, 2026, Dubai, United Arab Emirates}
\acmDOI{10.1145/3774904.3792940}
\acmISBN{979-8-4007-2307-0/2026/04}

\makeatletter
\gdef\@copyrightpermission{
  \begin{minipage}{0.2\columnwidth}
   \href{https://creativecommons.org/licenses/by/4.0/}{\includegraphics[width=0.90\textwidth]{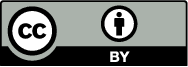}}
  \end{minipage}\hfill
  \begin{minipage}{0.8\columnwidth}
   \href{https://creativecommons.org/licenses/by/4.0/}{This work is licensed under a Creative Commons Attribution International 4.0 License.}
  \end{minipage}
  \vspace{5pt}
}
\makeatother

\usepackage{array}            
\usepackage{graphicx}         
\usepackage{subcaption}       

\usepackage{colortbl}         
\usepackage{soul}             

\usepackage{multirow}         
\usepackage{multicol}         
\usepackage{makecell}         
\usepackage[normalem]{ulem}

\usepackage{enumitem}         
\usepackage{pifont}           

\begin{document}

\title[Vertical Semi-Federated Learning for Efficient Online Advertising]{Vertical Semi-Federated Learning for Efficient Online Advertising}

\author{Wenjie Li}
\email{liwj20@mails.tsinghua.edu.cn}
\affiliation{
  \institution{Tsinghua Shenzhen International Graduate School, Tsinghua University}
  \city{Shenzhen}
  \country{China}
}
\affiliation{
  \institution{Meituan}
  \city{Shanghai}
  \country{China}
}

\author{Shu-Tao Xia}
\authornote{Corresponding author.}
\email{xiast@sz.tsinghua.edu.cn}
\affiliation{
  \institution{Tsinghua Shenzhen International Graduate School, Tsinghua University}
  \city{Shenzhen}
  \country{China}
}

\author{Jiangke Fan}
\email{811268232@qq.com}
\author{Teng Zhang}
\email{zhangteng09@meituan.com}
\author{Xingxing Wang}
\email{wangxingxing04@meituan.com}
\affiliation{
  \institution{Meituan}
  \city{Shanghai}
  \country{China}
}

\renewcommand{\shortauthors}{Wenjie Li et al.}

\newcommand{\eg}{\textit{e.g.}}
\newcommand{\cmark}{{\color{green}\ding{51}}}%
\newcommand{\xmark}{{\color{red}\ding{55}}}%

\newcommand{\myuparrow}{\contourlength{0.02em}\contour{black}{$\uparrow$}}
\newcommand{\myXm}{{\color{black}\xmark}}
\newcommand{\myCm}{{\color{black}\cmark}}
\newcommand{\XQL}[1]{\textcolor{blue}{#1}}
\newcommand{\myColor}[1]{{\color[HTML]{3166ff} #1}}

\begin{abstract}
  Traditional vertical federated learning schema suffers from two main issues: \textbf{\textit{1)}} \textit{restricted applicable scope to overlapped samples} and \textbf{\textit{2)}} \textit{high system challenge of real-time federated serving}, which limits its application to advertising systems. To this end, we advocate a new practical learning setting, \textbf{Semi-VFL} (Vertical Semi-Federated Learning), for real-world industrial applications, where the learned model retains sufficient advantages of federated learning while supporting independent local serving. To achieve this goal, we propose the carefully designed \textbf{J}oint \textbf{P}rivileged \textbf{L}earning framework (\textbf{JPL}) to \textbf{\textit{i)}} \textit{alleviate the absence of the passive party’s feature} with federated equivalence imitation and \textbf{\textit{ii)}} \textit{adapt to the heterogeneous full sample space} with cross-branch rank alignment. Extensive experiments conducted on real-world advertising datasets validate the effectiveness of our method over baseline methods.
\end{abstract}



\maketitle 

\section{Introduction}

\begin{table}[t]
\renewcommand{\arraystretch}{0.9}
\centering
\caption{\small The spectrum of different learning settings. Semi-VFL maximizes data usage while supporting local inference. (variables of unaligned samples are denoted by $u$, $A/B$ denotes for party field set.)}\label{tab:setting}
\begin{tabular}{c|c|c|c}
 \hline
 \textbf{Setting} & \textit{Local} & \textit{VFL} & \textit{Semi-VFL} \\
 \hline
 \textbf{Train\&Test Scope} & $N + N_u$ & $N$ & $N + N_u$ \\
 \textbf{Training Input} & $\mathbf{X}_A^u;\enspace \mathbf{X}_A$ & $[\mathbf{X}_A|\mathbf{X}_B]$ & $\mathbf{X}^{u}_A;\enspace [\mathbf{X}_A|\mathbf{X}_B]$ \\
 \textbf{Test Input} & $\mathbf{X}^u_A;\enspace \mathbf{X}_A$ & $[\mathbf{X}_A| \mathbf{X}_B]$ & $\mathbf{X}^u_A;\enspace \mathbf{X}_A$ \\
 \textbf{Local Inference} & \myCm & \myXm & \myCm \\
 \hline
\end{tabular}
\vspace{-1.5em}
\end{table}

Vertical Federated Learning (VFL) \cite{vfl_survey1} has emerged as a proactive solution \cite{wei2023fedads,ouyang2024fedud,li2024refer} for leveraging cross-platform user data to enhance recommendation and advertising performance, while preserving privacy and complying with regulations. Specifically, in a typical two-party VFL setting\cite{splitNNVP}, There is an \textit{{active party}} A who holds labels and part of the features and a \textit{{passive party}} B who provides additional features. They each privately hold a portion of the model and jointly train a federated model by communicating only hidden features and the corresponding gradients. In this way, Party A can leverage additional data fields from the B-side to improve its own task performance while preserving privacy.

Despite its promise, two main drawbacks hinders its deployment in practical scenarios. \textbf{(1) Restricted scope on overlapped data}: Vanilla VFL trains and serves only overlapped users, who typically constitute a small portion of the population. This limits information richness during training and prevents non-overlapped users from benefiting at all. As a result, the active party has little incentive to deploy a high-cost distributed learning system that yields limited returns, especially when a simpler local model can already serve all users more cost-efficiently. \textbf{(2) High-cost distributed serving}: The distributed nature of VFL introduces extra latency from cross-party data transmission and poses engineering challenges due to heterogeneous network and computational conditions. While such overhead is tolerable during offline training, it becomes nearly infeasible for online inference, where advertising systems demand both high throughput and strict real-time latency (millions of QPS and 10–100 ms per request~\cite{scale_rtb_icdm}). These constraints make VFL deployment in production prohibitively expensive.

These limitations reveal the need for a practical learning framework that can leverage all labeled data while supporting local inference for all users. We term this paradigm \textbf{Semi-VFL} (Vertical Semi-Federated Learning) to highlight its partially federated nature. Since ideal VFL is often impractical to deploy, the goal of Semi-VFL is to outperform vanilla local models. Notably, An effective Semi-VFL solution must address two key challenges: \textit{\textbf{\underline{(1)}} Alleviating field missing}: preserving and generalizing passive-party field knowledge from federated training to local inference is essential for maintaining the advantages of VFL; \textit{\textbf{\underline{(2)}} Adapting to the heterogeneous full sample space}: effectively combining the imbalanced field information of overlapped and non-overlapped samples is crucial for achieving consistent gains across all users. Several recent studies explore integrating unaligned data or imputing B-side missing fields in the latent space \cite{ouyang2024fedud, kang2022fedcvt}, but they mainly aim to improve federated performance rather than support local inference. A few distillation-based approaches \cite{pri_fed, vfedssd} do produce local models compatible with Semi-VFL, yet they rely on direct logit distillation and lacks comprehensive cross-party knowledge transfer.

To tackle these problems, we propose a \textbf{J}oint \textbf{P}rivileged \textbf{L}earning (JPL) framework that preserves the strengths of both local and federated modeling. JPL adopts a two-branch architecture: one branch captures local-only inductive bias from active-party fields, while the other distills federated knowledge. In the federated branch, we perform \textit{\textbf{federated equivalence imitation}} by learning an A-to-B transformation with various imitation regularization, explicitly alleviating B-side field missing. In addition, we introduce a \textit{\textbf{cross-head rank alignment}} loss and ensemble the outputs of both branches to effectively integrate non-federated and federated inductive biases. In summary, our contributions are threefold: 
\begin{itemize}[label=$\clubsuit$, leftmargin=*, itemsep=0em, parsep=0em, topsep=0.2em]
\item We identify the \textit{vertical semi-federated learning} setting for advertising systems and highlight its practical importance in real-world industrial deployments.
\item We introduce \textbf{JPL}, a distillation-based Semi-VFL framework that integrates federated equivalence imitation and cross-head rank alignment to jointly enhance cross-party knowledge transfer under the constraint of local inference.
\item We conduct extensive experiments on benchmark datasets and demonstrate that JPL consistently outperforms strong baselines across all evaluation settings.
\end{itemize}

\section{Method}\label{sec:method}

\begin{figure}[t]
\centering
\includegraphics[width=\linewidth]{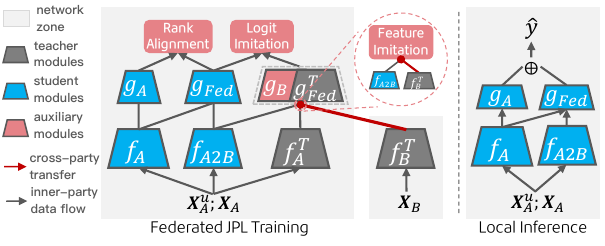}
\caption{\small The overview of JPL.} \label{fig:jpl}
\vspace{-2em}
\end{figure}

Figure \ref{fig:jpl} presents an overview of JPL, featuring two key aspects: \textbf{(1) Two-stage distillation pipeline}. A student model is trained over the full sample space under the guidance of a federated teacher model pre-trained on the overlapped subset. The resulting student can then perform inference independently without cross-party communication. \textbf{(2) Complementary two-branch architecture}. The student model consists of a local branch that learns inductive bias from the A-field’s full sample space and a federated branch that distills knowledge from the teacher built on the full feature space. These branches complement each other through implicit knowledge transfer via a shared A-field encoder and unified encoding flow for both overlapped and non-overlapped samples. To preserve their respective strengths, JPL applies tailored regularizations: \textbf{\textit{federated equivalence imitation}} at both logit and feature levels, paired with a \textbf{\textit{cross-head rank alignment}} mechanism that effectively fuses predictions from the two branches.


\subsection{Federated Equivalence Imitation}
In the federated branch, an A-to-B transformation encoder $f_{A2B}(\cdot)$ is introduced to learn cross-party feature mappings. It is jointly optimized for {\textit{logit imitation}} and {\textit{feature imitation}}, to achieve discriminative and representational equivalence, respectively.

\subsubsection{\textbf{Logit Imitation}} For the non-overlapped data $\mathbf{x}^u_A$, we denote the imitated feature as $\tilde{\mathbf{u}}_B = f_{A2B}(\mathbf{x}^u_A)$. We enforce its discriminative capability on both the full-side and the B-side fields:
\begin{align}
    \mathcal{L}^{ce}_{li}(\mathbf{x}^u_A) = CE(y, g^T_{Fed}(f^T_A(\mathbf{x}^u_A), \tilde{\mathbf{u}}_B)) + CE(y, g_B(\tilde{\mathbf{u}}_B)),
\end{align}
where $CE$ is the binary cross-entropy loss, $g^T_{Fed}(\cdot)$ and $f^T_A(\cdot)$ respectively denote the frozen classifier and A-side encoder from the teacher, and $g_B$ is a learnable auxiliary classifier held by the student to learn the B-side inductive bias. Unless otherwise indicated by the superscript $^T$, all modules refer to the student model.

For overlapped samples, the imitated and ground-truth B-side features are denoted as $\tilde{\mathbf{h}}_B = f_{A2B}(\mathbf{x}_A)$ and $\mathbf{h}_B = f_B(\mathbf{x}_B)$. In addition to the supervised signal used for non-overlapped data, the availability of $\mathbf{x}_B$ enables an extra logit imitation constraint:
\begin{align}
    \mathcal{L}^{kd}_{li}(\mathbf{x}_A) = &~KL[g^T_{Fed}(\mathbf{h}^T_A, \tilde{\mathbf{h}}_B) \| \hat{y}^T] +
    KL[g_B(\mathbf{h}^T_B) \| g_B(\tilde{\mathbf{h}}_B))],
\end{align}
where $\mathbf{h}^T_A = f^T_A(\mathbf{x}_A)$ and $\mathbf{h}^T_B = f^T_B(\mathbf{x}_B)$ are teacher encodings, and $KL$ denotes the KL-divergence. The first term aligns $\tilde{\mathbf{h}}_B$ with $\mathbf{h}_B$ in the teacher’s prediction space, while the second enforces consistency in the student’s B-side classifier. Finally, we get overlapped samples' overall logit imitation loss $\mathcal{L}_{li}(\mathbf{x}_A) = \mathcal{L}^{ce}_{li}(\mathbf{x}_A) + \mathcal{L}^{kd}_{li}(\mathbf{x}_A)$.

\subsubsection{\textbf{Feature Imitation}} 
Due to the intrinsic difference between field sets and the difficulty of fully reconstructing B-side features, we adopt a relative reconstruction strategy based on similarity consistency rather than direct feature recovery. Specifically, letting $\tilde{\mathbf{H}}_B = f_{A2B}(\mathbf{X}_A)$ and $\mathbf{H}_B = f_B(\mathbf{X}_B)$ denote the imitated and ground-truth B-side features of a batch, the feature imitation loss for overlapped samples is defined as:
\begin{align}
    \mathbf{C} &= \tilde{\mathbf{H}}_B\odot (\mathbf{H}_B^T)^\top - \mathbf{H}_B^T \odot(\mathbf{H}_B^T)^\top, \\
    \mathcal{L}^B_{fi}(\mathbf{X}_A) &= \frac{1}{N}\|\text{diag}(\mathbf{C})\|_2^2 + \frac{1}{N(N-1)}\|\mathbf{C} - \text{diag}(\mathbf{C})\|_F^2,
\end{align}
where $\odot$ denotes matrix multiplication with row-wise $L_2$ normalization. $\text{diag}(\cdot)$ denotes the diagonal vector. We balance the impact of different pairs by normalizing diagonal and off-diagonal terms: intuitively, paired student–teacher similarities (diagonal) are fewer but more important, while cross-sample similarities (off-diagonal) are more abundant but less critical.

For non-overlapped samples, no ground-truth B-side features are available for direct constraints. We therefore use the aligned data as an anchor and minimize the cross-field-space similarity gap, assuming that sample similarities in field spaces A and B should be consistent. Specifically:
\begin{equation}
    \mathcal{L}^{AB}_{fi}(\mathbf{X}^u_A) = \|f^T_A(\mathbf{X}^u_A)\odot\mathbf{H}_A^T - f_{A2B}(\mathbf{X}^u_A)\odot\mathbf{H}_B^T\|_F^2.
\end{equation}
Here we use the teacher’s features as anchors for stable supervision.

\subsection{Cross-Head Rank Alignment}
To integrate the strengths of both branches, we propose a \textbf{Privileged Ranking Consistency} (PRC) loss that enforces rank-level alignment between the prediction heads. PRC aims to maximize pairwise ranking consistency under privileged supervision. Specifically, we represent the ranking of a batch of predictions $\hat{\mathbf{y}}$ with a \textit{partial order matrix} (POM) $\mathbf{R}$, where each element $\mathbf{R}_{ij} = \sigma(\hat{y}_i - \hat{y}_j)$ denotes the probability $p(\hat{y}_i > \hat{y}_j)$ modeled by a logistic function $\sigma(\cdot)$. The POM is organized according to the label space:
\begin{equation}
    \mathbf{R} = 
    \begin{bmatrix}
        \mathbf{R}^{++} & \mathbf{R}^{+-} \\
        (\mathbf{R}^{+-})^\top & \mathbf{R}^{--}
    \end{bmatrix},
\end{equation}
where the superscripts “+” and “–” indicate label categories of the compared pairs, e.g., $\mathbf{R}^{+-} = \{ r_{ij} \mid i \in \mathcal{Y}^+, j \in \mathcal{Y}^- \}$. The PRC loss for overlapped samples is formulated as:
\begin{align}
    \mathcal{L}^{A\leftarrow F}_{rank} &=
    \frac{\|\mathbf{R}^{++}_A - s(\mathbf{R}^{++}_{Fed})\|_F}{\|s(\mathbf{R}^{++}_{Fed})\|_F}
    + \frac{\|\mathbf{R}^{--}_A - s(\mathbf{R}^{--}_{Fed})\|_F}{\|s(\mathbf{R}^{--}_{Fed})\|_F}
    - \|\mathbf{R}^{+-}_A\|_F, \notag
\end{align}
where $s(\cdot)$ denotes the stop-gradient operation, and the subscripts “A” and “Fed” refer to the local and federated heads, respectively. The first two terms reduce intra-class ranking discrepancies between the heads (with frozen gradients on the federated head) and normalize by label scale, while the last term fits the fixed cross-class rank. For non-overlapped data, the PRC loss follows the same formulation but reverses the alignment direction as $\mathcal{L}^{F\leftarrow A}_{rank}$. Finally, the predictions from both heads are fused by averaging logits $\hat y = \sigma(\hat y_A/2+\hat y_{Fed}/2)$.

Putting all together, we get the overall learning objective:
\begin{align}
    \mathcal{L}(\mathbf{X}^u_A) &= \mathcal{L}^{F\leftarrow A}_{rank} + \beta\cdot\mathcal{L}^{AB}_{fi} + \mathcal{L}^{ce}_{li}\\
    \mathcal{L}(\mathbf{X}_A) &= \mathcal{L}^{A\leftarrow F}_{rank} + \beta\cdot\mathcal{L}^B_{fi}+ \mathcal{L}^{ce}_{li} + \mathcal{L}^{kd}_{li} 
\end{align}
where $\beta$ is set to control the impact of feature imitation.

\section{Experiments}\label{sec:exp}

\renewcommand{\arraystretch}{0.8}
\begin{table*}[t]
\centering
\caption{\small The overall results on all sample subsets. Typically, an AUC increase of 0.001 can be considered a significant in CTR Prediction\cite{zhou2018din}.}
\label{tab:main_exp}
\resizebox{0.9\textwidth}{!}{
\begin{tabular}{r|cc|cc|cc|cc|cc|cc}
\toprule
\multicolumn{1}{l|}{} & \multicolumn{6}{c|}{Avazu}                             & \multicolumn{6}{c}{Criteo}                             \\ \midrule
\multicolumn{1}{l|}{} &
  \multicolumn{2}{c|}{overall} &
  \multicolumn{2}{c|}{unaligned} &
  \multicolumn{2}{c|}{aligned} &
  \multicolumn{2}{c|}{overall} &
  \multicolumn{2}{c|}{unaligned} &
  \multicolumn{2}{c}{aligned} \\ \midrule
Method                & AUC$\uparrow$& Logloss$\downarrow$& AUC$\uparrow$& Logloss$\downarrow$& AUC$\uparrow$& Logloss$\downarrow$& AUC$\uparrow$& Logloss$\downarrow$& AUC$\uparrow$& Logloss$\downarrow$& AUC$\uparrow$& Logloss \\ \midrule

Fed &
  \textbackslash{} &
  \textbackslash{} &
  \textbackslash{} &
  \textbackslash{} &
  0.7082 &
  0.3768 &
  \textbackslash{} &
  \textbackslash{} &
  \textbackslash{} &
  \textbackslash{} &
  0.7844 &
  0.4644 \\

Local                 & 0.7082 & {0.3654}  & {0.7172} & {0.3505}  & 0.6992 & 0.3803  
                     & {0.7750} & 0.4684  & 0.7768 & {0.2332}  & 0.7734 & 0.2352  \\
FPD                   & 0.7103 & \uwave{0.3611}  & 0.7166 & \uwave{0.3476} & 0.7037 & 0.3746  
                     & \uwave{0.7764} & \uwave{0.4672}  & \uwave{0.7785} & \uwave{0.2326}  & \uwave{0.7746} & \uwave{0.2347}  \\

FedUD*                 & 0.7055 & 0.3708  & 0.7129 & 0.3587  & 0.6982 & 0.3829  
                     & 0.7711 & 0.4712  & 0.7728 & 0.4695  & 0.7694 & 0.4729  \\ 
FedCVT*                 & \uwave{0.7145} & 0.3616  & \uwave{0.7195} & 0.3498  & \uwave{0.7089} & \uwave{0.3735}  
                       & 0.7653 & 0.4776  & 0.7674 & 0.4752  & 0.7634 & 0.4800  \\
\midrule
\rowcolor{green!10}
\textbf{JPL}                   & \textbf{0.7207} & \textbf{0.3597}  
                              & \textbf{0.7279} & \textbf{0.3468}  
                              & \textbf{0.7127} & \textbf{0.3725}  
                              & \textbf{0.7775} & \textbf{0.4667}  
                              & \textbf{0.7794} & \textbf{0.2324}  
                              & \textbf{0.7758} & \textbf{0.2343}  \\ 
\bottomrule
\end{tabular}
\vspace{-2em}
}
\end{table*}

\subsection{Experimental Settings}

\subsubsection{\textbf{Datasets}}
We evaluate our method on two widely used public CTR prediction datasets, \textit{Criteo} and \textit{Avazu} under a VFL data partition setting as in ~\cite{vfedssd}. Specifically, for \textit{Avazu}, both parties hold 11 feature fields, while for \textit{Criteo}, the active and passive parties have 19 and 20 fields, respectively. We respectively construct 6, 1 and 1 million samples for training validation and testing. Following \cite{kang2022fedcvt}, we select earlier half data to construct non-overlapped samples. 

\subsubsection{\textbf{Baselines and Evaluation Setting}}

We compare our approach against four baselines: \textbf{(1) Local.} A local model is trained solely on the active party’s features. A qualified Semi-VFL method is expected to outperform this baseline. \textbf{(2) FPD}~\cite{pri_fed,vfedssd}. These pioneering works first train a federated model on overlapped samples and then distill logit knowledge into a local model, thus we collectively refer to them as FPD (federated privileged distillation) methods. In our implementation, the training objective is defined as $\mathcal{L}_{FPD}= \mathcal{L}^n_{CE} + (1-\alpha)\mathcal{L}^o_{CE} + \alpha\mathcal{L}^o_{KL}$, where $\alpha$ denotes the distillation strength. \textbf{(3) FedUD*}~\cite{ouyang2024fedud}. FedUD leverages unaligned samples by generating passive-side representations using a pre-trained cross-party feature transfer encoder. Since the original method does not support local inference, we modify it to employ the transfer encoder for both aligned and unaligned samples. \textbf{(4) FedCVT*}~\cite{kang2022fedcvt}. This method leverages unaligned samples during training by imputing B-side hidden representations from aligned samples’ B-side features using its similarity-based mechanism. It effectively learns implicit A-to-B transferable knowledge that can exploit unaligned samples. However, this imputation is not applicable under local inference. Although originally designed for the full VFL scenario, we modify it to use only classifier A’s output, making it a comparable baseline for local inference. Following~\cite{wei2023fedads}, we implement an efficient, fast version of this method. We report AUC and logloss results of all baselines evaluated on All sample group under the \textit{\textbf{local inference mode}}, where only local fields are available during inference.

\subsubsection{\textbf{Implementation Details}}
For the model structure, as our method is orthogonal to architecture design, we use a two 512-unit dnn layer as bottom model and a single output layer as bootom model by default. We report extra results on other archetectures in Table~\ref{tab:backbone_exp}. For training, we use the Adam optimizer with $L_2$ regularization. In all experiments, the default learning rate and L2 regularization strength are both 0.001, and the batch size is $10K$. The distillation strength for FPD is chosen from $\alpha \in {0.1, 0.5, 1}$. For FedCVT, the regularization strength of the feature imputation module is selected from ${0.1, 0.5, 1}$, and the maximum auxiliary aligned data queue size is 20K. For FedUD, both the regularization ratio of the transfer encoder and the loss weight of unaligned samples are searched over ${0.1, 0.5, 1}$. For JPL, the feature imitation loss ratios $\beta$ are set according to their magnitudes (\textit{e.g.}, $\{0.1, 0.5, 1\}$ for $\mathcal{L}^{B}_{fi}$ and $\{100, 500, 1000\}$ for $\mathcal{L}^{AB}_{fi}$). The same teacher model is used for all distillation-based methods. We tune hyper-parameters for all methods and report the best result.


\subsection{Results}
\subsubsection{\textbf{Overall Performance}}
The main results in Table~\ref{tab:main_exp} lead to three key observations. 
\textit{\underline{(1)}} \textbf{JPL achieves the strongest overall performance} on both Avazu and Criteo, across all settings (overall, unaligned, and aligned), it can improves the Local model up to \textit{+0.0110\,$\sim$\,0.0025} on overall AUC respectively. on Avazu dataset, Its performance on aligned dataset can even outperform the federated teacher model, confirming its capability of integrating advantage from both sample space and field space. \textit{\underline{(2)}} \textbf{Distillation-based methods} (FPD and JPL) consistently outperform Local on all subsets, showing that directly transferring the teacher’s knowledge into independent local model yields stable and reliable gains. \textit{\underline{(3)}} \textbf{Adapted federated methods} (FedUD and FedCVT) generally show unsatisfactory performance, with only occasional improvements (\eg, FedCVT on Avazu reaches the second-best overall AUC), indicating unstable adaptation to heterogeneous sample distributions. This is reasonable because these methods are not tailored for Semi-VFL. Although they consider transferring federated feature knowledge during training, they neither fully support cross-party knowledge transfer based solely on A-side features nor effectively combine information from the full sample space.

\begin{table}[t]
\centering
\vspace{-1em}
\caption{\small Component Ablation Study. Removing any component leads to a performance drop, indicating each module' necessity.}
\label{tab:ablation}
\resizebox{\columnwidth}{!}{%
\begin{tabular}{@{}r|ccc|ccc@{}}
\toprule
\multicolumn{1}{c|}{} & \multicolumn{3}{c|}{Avazu}                                         & \multicolumn{3}{c}{Criteo}                                         \\ \midrule
Method & \multicolumn{1}{c|}{overall} & \multicolumn{1}{c|}{unaligned} & aligned & \multicolumn{1}{c|}{overall} & \multicolumn{1}{c|}{unaligned} & aligned \\ \midrule
w/o logit imit.   & \multicolumn{1}{c|}{0.6996} & \multicolumn{1}{c|}{0.7149} & 0.6855 & \multicolumn{1}{c|}{0.7659} & \multicolumn{1}{c|}{0.7683} & 0.7637 \\
w/o feat imit.   & \multicolumn{1}{c|}{0.7149} & \multicolumn{1}{c|}{0.7230} & 0.7067 & \multicolumn{1}{c|}{0.7750} & \multicolumn{1}{c|}{0.7772} & 0.7729 \\
w/o rank align & \multicolumn{1}{c|}{0.7068} & \multicolumn{1}{c|}{0.7161} & 0.6972 & \multicolumn{1}{c|}{0.7757} & \multicolumn{1}{c|}{0.7777} & 0.7737\\
\bottomrule
\end{tabular}%
\vspace{-1.5em}
}
\end{table}

\subsubsection{\textbf{Ablation Study}}
To evaluate the contribution of each component in JPL, we conduct an ablation study reported in Table~\ref{tab:ablation}. The three rows present the performance after individually removing each module.
\textit{(1)} The results show that all components contribute positively, as removing any of them leads to a noticeable performance drop compared with the overall results in Table~\ref{tab:main_exp}.
\textit{(2)} Removing the logit imitation loss produces the largest degradation, followed by rank alignment and then feature imitation. This is expected since logit imitation provides the most direct supervisory signal by aligning both labels and teacher logits, while feature imitation plays a more auxiliary, intermediate role. The clear benefit of rank alignment suggests that the two heads capture complementary information, allowing mutual ranking guidance and logit ensembling to enhance overall performance.

\subsubsection{\textbf{Impact of Data Volume and Field Amount}}
\begin{figure}[t]
    \centering
    \vspace{-1em}
    \includegraphics[width=0.24\linewidth]{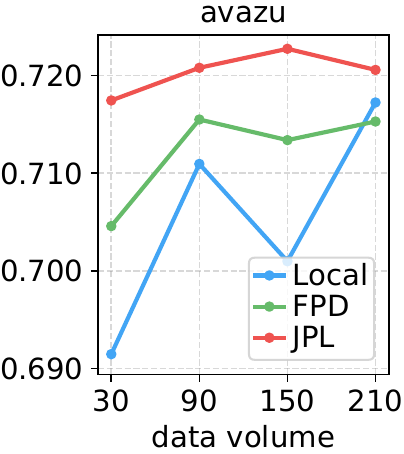}
    \includegraphics[width=0.24\linewidth]{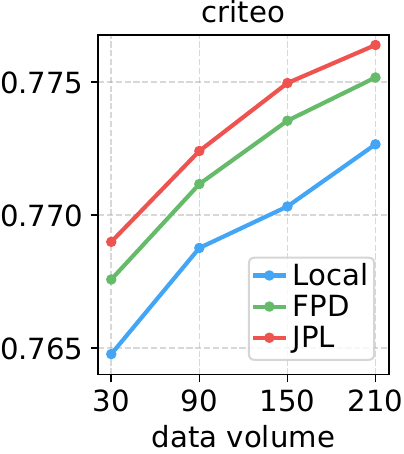}
    \includegraphics[width=0.24\linewidth]{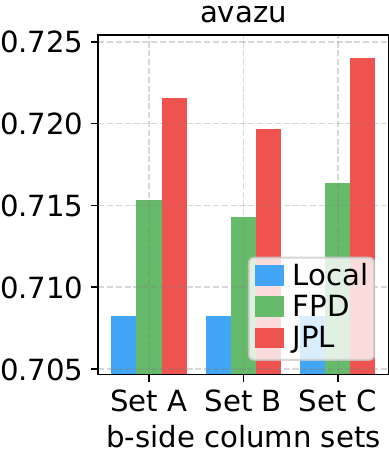}
    \includegraphics[width=0.24\linewidth]{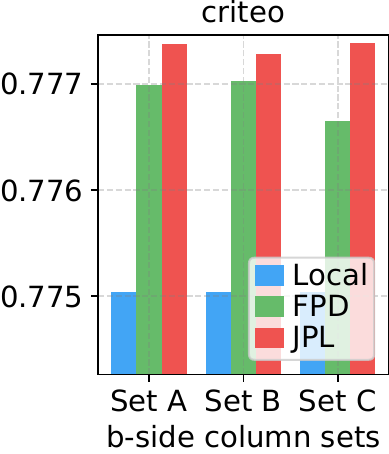}
    \caption{\small Impact of non-overlapped data volume (left) and the passive party's field set (right). JPL maintains consistent advantage.}\label{fig:un_volume_exp}
    \vspace{-1.5em}
\end{figure}

As shown in Figure \ref{fig:un_volume_exp}(left), we conducted experiments on datasets with decreasing proportions of unaligned samples: 70\%, 50\%, 30\%, and 10\%. It shows that incorporating more unaligned data generally leads to a performance increase, while the advantage of distillation remains consistent across different data volumes. While in Figure \ref{fig:un_volume_exp}(right), we vary the field sets by using 25\%, 50\%, and 75\% of the fields (Sets A, B, and C) to assess robustness across field configurations. Note that due to differences in field importance and vocabulary distributions, more fields do not necessarily implies better input information. The results show that our method consistently outperforms both Local and FPD across all configurations. Since the Local model is unaffected by this operation, its performance bars remain identical.

\subsubsection{\textbf{Effectiveness across Different Architectures}}
To evaluate the generalizability of our framework to different backbone architectures, we design two variants of Local, FPD and JPL based on DeepFM \cite{guo2017deepfm} and DCN \cite{wang2017DCN}. These architectures are applied to the student’s local bottom model, and new students are trained while keeping the teacher model unchanged. As shown in Table \ref{tab:backbone_exp}, JPL consistently demonstrates superior performance, validating the effectiveness of our framework under heterogeneous teacher–student setups and across various local model architectures.

\begin{table}[t]
\centering
\vspace{-1em}
\caption{\small Architectures Compatibility Study. The consistent advantage of JPL confirms its generalizability and effectiveness when integrated with diverse network architectures.} \label{tab:backbone_exp}
\resizebox{\columnwidth}{!}{

\begin{tabular}{@{}cc|ccc|ccc@{}} 
\toprule
&     & \multicolumn{3}{c|}{Avazu} & \multicolumn{3}{c}{Criteo} \\ \midrule
Backbone & Method & {overall} & {unaligned} & aligned & {overall} & {unaligned} & aligned \\
\midrule
\multirow{3}{*}{DCN} & Local & {0.7088} & {0.7168} & 0.7012 & {0.7748} & {0.7767} & 0.7731 \\
& FPD & {0.7133} & {0.7218} & 0.7039 & {0.7772} & {0.7793} & 0.7752 \\
& \cellcolor{green!10} \textbf{JPL} & \cellcolor{green!10}{\textbf{0.7190}} & \cellcolor{green!10} {\textbf{0.7270}} & \cellcolor{green!10} \textbf{0.7105} & \cellcolor{green!10} {\textbf{0.7780}} & \cellcolor{green!10} {\textbf{0.7800}} & \cellcolor{green!10} \textbf{0.7760} \\
\midrule
\multirow{3}{*}{DeepFM} & Local & {0.7118} & {0.7184} & 0.7057 & {0.7764} & {0.7784} & 0.7746 \\
& FPD & {0.7126} & {0.7195} & 0.7054 & {0.7777} & {0.7795} & 0.7760 \\
& \cellcolor{green!10}\textbf{JPL} & \cellcolor{green!10} {\textbf{0.7220}} & \cellcolor{green!10} {\textbf{0.7296}} & \cellcolor{green!10} \textbf{0.7138} & \cellcolor{green!10} {\textbf{0.7782}} & \cellcolor{green!10} {\textbf{0.7801}} & \cellcolor{green!10} \textbf{0.7765} \\
\bottomrule

\end{tabular}
\vspace{-1.5em}
}
\end{table}

\section{Conclusion}
This paper presents a Joint Privileged Learning (JPL) framework for vertical semi-federated advertising. JPL introduces federated equivalence imitation to mitigate field-missing issues and cross-head rank alignment to enable efficient knowledge fusion across full-sample and full-field spaces. Extensive experiments on two widely used CTR datasets validated its effectiveness and rationality.

\begin{acks}
    This work is supported in part by the National Natural Science Foundation of China under Grant 62571298. Thanks to Tencent Rhino-bird Elite Talent Program for the early start of this project.
\end{acks}

\bibliographystyle{ACM-Reference-Format}
\bibliography{main}

\end{document}